\documentclass[preprint,12pt]{elsarticle}

\usepackage{amssymb}
\usepackage{amsmath}
\usepackage{url}

\begin{document}
\begin{frontmatter}

\title{\textbf{Dataset} $\mid$ \textbf{Mindset = Explainable AI} $\mid$ \textbf{Interpretable AI}}

\author{$Caesar Wu^a,  Rajkumar Buyya^b, Yuan Fang Li^c, Pascal Bouvry^a$} 

\affiliation{organization={SnT, Department of Computer Science, Faculty of Science Technology and Medicine (FSTM), University of Luxembourg}, 
            addressline= {6 Avenue de la Fonte}, 
            city={Belval Campus},
            postcode={4364}, 
            state={Esch-sur-Alzette},
            country={Luxembourg}}
           
\affiliation{organization={CLOUDS lab, School of Computing and Information Systems, The University of Melbourne}, 
            addressline= {700 Swanston St}, 
            city={Carlton},
            postcode={3053}, 
            state={Victoria},
            country={Australia}}
           
\affiliation{organization={Computer Science, Faculty of Information Technology, Monash University}, 
            addressline= {20 Exhibition Walk}, 
            city={Clayton},
            postcode={3800}, 
            state={Victoria},
            country={Australia}}

\begin{abstract}
We often use "explainable" Artificial Intelligence (XAI)" and "interpretable AI (IAI)" interchangeably when we apply various XAI tools for a given dataset to explain the reasons that underpin machine learning (ML) outputs. However, these notions can sometimes be confusing because interpretation often has a subjective connotation, while explanations lean towards objective facts. We argue that XAI is a subset of IAI. The concept of IAI is beyond the sphere of a dataset. It includes the domain of a mindset. At the core of this ambiguity is the duality of reasons, in which we can reason either outwards or inwards. When directed outwards, we want the reasons to make sense through the laws of nature. When turned inwards, we want the reasons to be happy, guided by the laws of the heart. While XAI and IAI share reason as the common notion for the goal of transparency, clarity, fairness, reliability, and accountability in the context of ethical AI and trustworthy AI (TAI), their differences lie in that XAI emphasizes the post-hoc analysis of a dataset, and IAI requires a priori mindset of abstraction. This hypothesis can be proved by empirical experiments based on an open dataset and harnessed by High-Performance Computing (HPC). The demarcation of XAI and IAI is indispensable because it would be impossible to determine regulatory policies for many AI applications, especially in healthcare, human resources, banking, and finance. We aim to clarify these notions and lay the foundation of XAI, IAI, EAI, and TAI for many practitioners and policymakers in future AI applications and research. 
\end{abstract}


\end{frontmatter}

\section{Introduction}
\label{sec1}
The notion of eXplainable Artificial Intelligence (XAI) suggests the ability to clarify AI. It offers reasons, evidence, and contexts for the Artificial Intelligence/Machine Learning (AI/ML) results. It answers questions of "why" and "how". "Why" offers reasons, and "how" explains how the AI arrived at the result supported by reasons. It is a process of making sense of the AI/ML outputs. 

On the other hand, Interpretable AI (IAI) also provides reasons, but it asks the question, "Is the reason reasonable?" If we consider IAI to be a high-level abstraction of XAI, IAI is the meta-XAI or criteria of XAI for satisfaction. The IAI process requires our mindset. Molnar\cite{Mohar33} argued that our mindset is a perspective of the world when we construct a learning model. In computer programming, "interpretation" means interactively converting a high-level language into machine codes line-by-line. A programming language is a form of communication between programmers and other people. It also helps programmers organize and describe ideas for a computer. Similarly, IAI can be considered a form of communication among humans, while XAI is an engagement process of our mindset for a given dataset within a particular problem context.

Guidotti et al.\cite{Guidotti1} argued that IAI is needed for AI models, while XAI aims for prediction results. Miller\cite{Miller2} attempted to define XAI from different perspectives, including philosophy, cognitive psychology/science, and social psychology. Miller did not differentiate between interpretability (XAI) and explainability (IAI). Molnar\cite{Molnar16} also prefers to use both terms interchangeably but differentiates between explainability/interpretability (degree to human understanding) and explanation(predictions). Lipton \cite{Lipton3} defines "interpretability" from an ML algorithm perspective but finds the term is slippery. He argued that today's predictive (AI/ML) models are incapable of reason at all. Nevertheless, Lipton proposed two approaches to understanding XAI: the intrinsic (thing-in-itself) and the "post-hoc" methods. He concludes that understanding "thing in itself" is to interpret, while "post-hoc" is to explain. 

Benois-Pineau et al. \cite{Benois4} define "explaining" as the process of computing, while "interpreting" is the process of assigning meaning to the explanation, which sometimes also applies to the model's representation. They infer that interpretation involves human decision-making for the model's representation.

Likewise, Brain and Cotton \cite{Biran5} argue that the concept of explainability is closely related to interpretability because if ML models are interpretable, their operations can be understood by a human through a process of explanation. They claim that the interpretable ML models can intrinsically be explained through reasoning. Following a similar logic, Burkart and Huber \cite{Burkart6} describe "interpretation" as interpreting ML models, while "explanation" means providing reasons.

Although many researchers have made great contributions to differentiate and articulate the meaning of interpretable AI (IAI) and eXplainable AI (XAI) \cite{DARPA7}, confusion has still remained. As Zhong and Negre \cite{Zhong8} argued, the AI research community failed to reach a consensus for a strict definition of interpretability and explainability. Many ambiguities exist not only in how to define the terms but also in how to evaluate the ML models. They concluded that many definitions are often cast out arbitrarily or subjectively.

\begin{figure}[htb]
    \centering
    \includegraphics[width=0.8\linewidth]{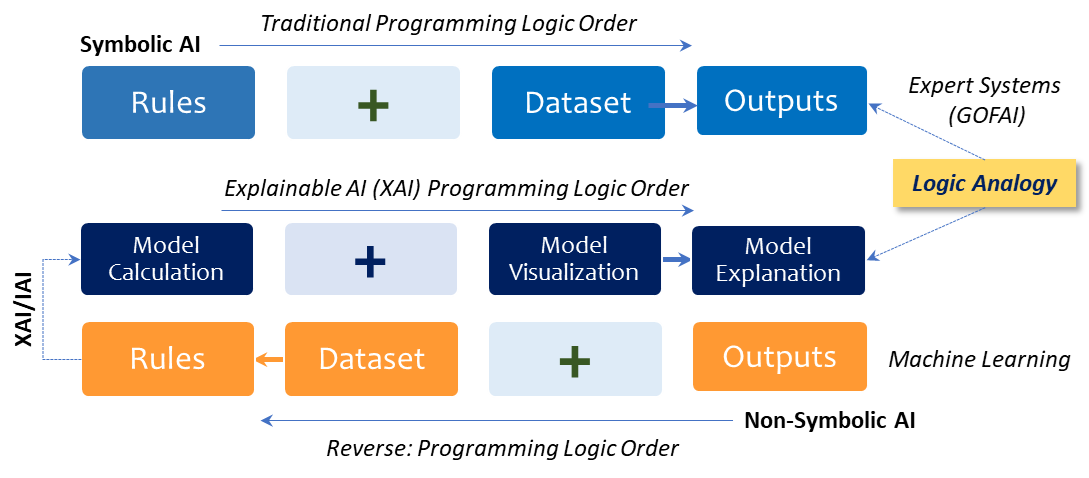}
    \caption{Reverse Programming Logic}
    \label{fig:1}
    \vspace{-0.3 cm}
\end{figure}

The fundamental issue is that the primary question remains unclear or has not been asked: "If the ML model can provide a good prediction outcome, why do we still need an explanation or interpretation?" What is the essential difference between symbolic AI or Good Old Fashioned Artificial Intelligence (GOFAI) and modern AI/ML? The answer lies in the ML process, which is reversed programming logic.(See Figure.~\ref{fig:1})

\begin{figure}[htb]
    \centering
    \includegraphics[width=0.8 \linewidth]{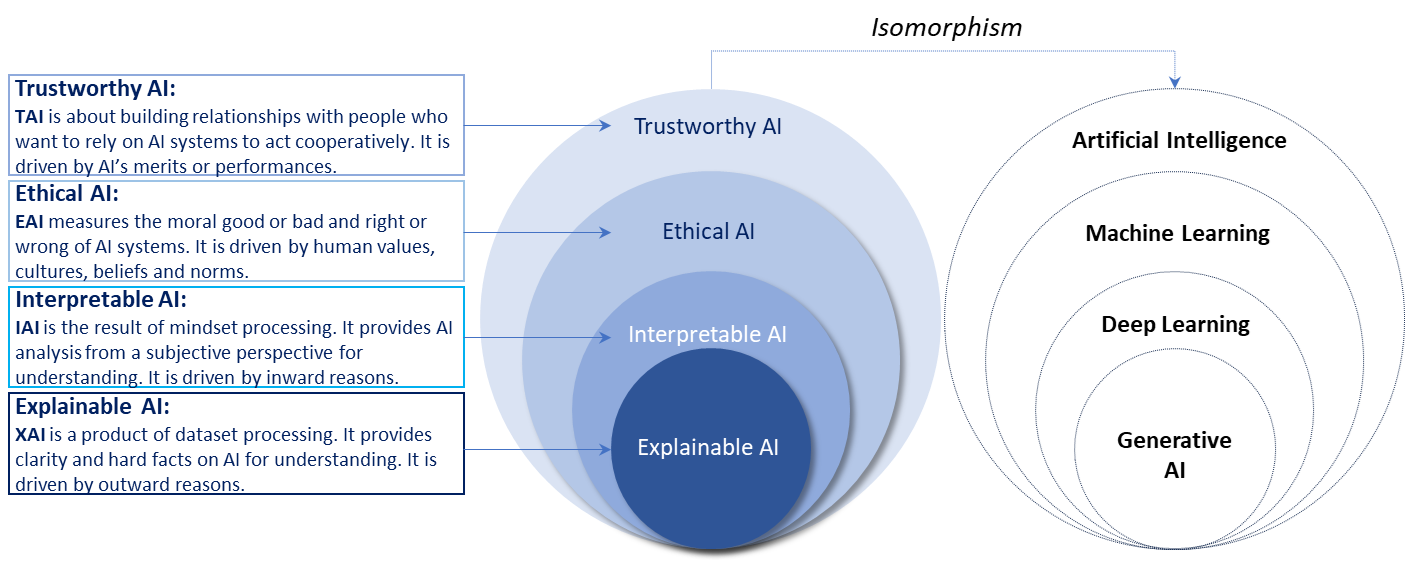}
    \caption{Relationship of TAI, EAI, IAI, and XAI}
    \label{fig:17}
    \vspace{-0.3 cm}
\end{figure}

This reversed programming logic drives the demand for XAI and IAI because the data patterns generate a set of rules and prediction models. Consequently, the ML result could become unexplainable. Therefore, we long for XAI and IAI. To some extent, the essence of the XAI/IAI process is to reverse the reserved programming logic order. We can draw the logic analogy between XAI/IAI and traditional programming logic order. Implicitly, the process of explanation and interpretation shares the common notion (reason) with a special characteristic of duality. This is why people use both terms interchangeably. From a hierarchical perspective, XAI is a subset of IAI. IAI is a subset of EAI, and EAI is a subset of TAI by drawing an isomorphism \cite{Popova38}from the relation of AI, ML, Deep Learning (DL), and Generative AI (Refer to Figure ~\ref{fig:17}) because we need XAI to reverse "reversed programming logic", IAI to understand XAI's process, EAI to satisfy IAI's reasons or values, and TAI to rely on EAI criteria.

\begin{figure}[htb]
    \centering
    \includegraphics[width=0.9 \linewidth]{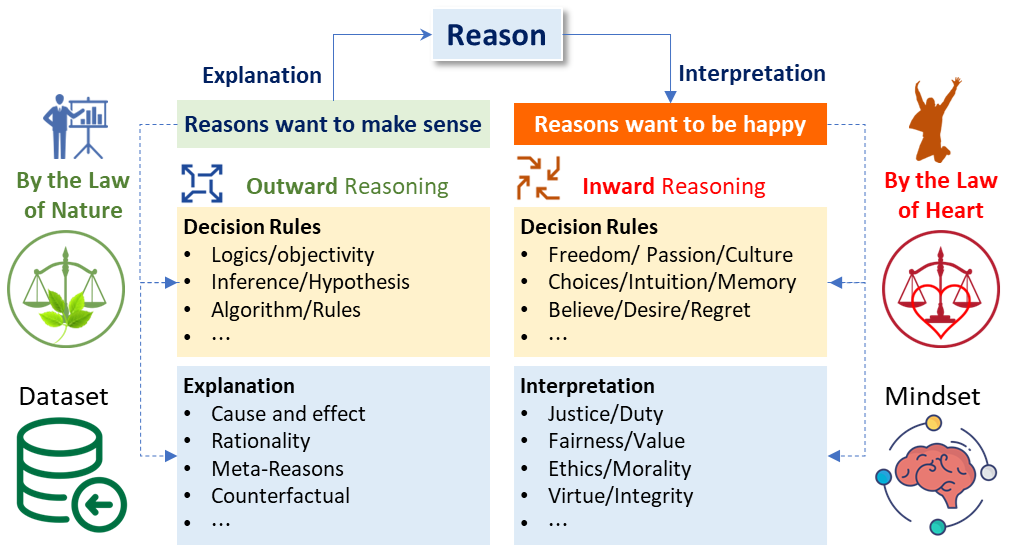}
    \caption{The Duality of Reason for Explanation and Interpretation}
    \label{fig:2}
    \vspace{-0.5 cm}
\end{figure}

From a reasoning perspective, we can find that when reason is faced outward, it sees logic, objectivity, inference, rules, and algorithms. Reasons (we) want to make sense. It explains cause-and-effect, deductive, inductive, analogical, correlational, probabilistic, and counterfactual inference. The law of nature governs the XAI. However, when reasons (we) are faced inward, they do not see all logical objects in an array. Instead, they see "a blooming and buzzing confusion" with half thoughts, fuzzy memories, and some unpleasant regrets.\cite{Oneill9}. In short, reasons want to be happy. The inward reasons reveal our mind, internal freedom, passion, choice, intuitions, desires, and beliefs. They give rise to fairness, ethics, morals, and justice \cite{Logins10}. They derive from the law of the heart and eventually become the law of ethics and morality (See Figure ~\ref{fig:2}). In other words, XAI and IAI are based on different decision rules or frames of mind. 

Hence, we articulate the demarcation between XAI and IAI. This demarcation can be demonstrated through a series of empirical experiments by adopting widely recognized XAI techniques \cite{Greenwell34, Dandl35, bradley36}. We aim to draw many practitioners' and policymakers' attention to this essential difference and lay the groundwork for future XAI, IAI, EAI, and Trustworthy AI (TAI) research and AI applications. By doing so, we made the following contributions:

\begin{itemize}

    \item We define the duality of reason as the demarcation between XAI and IAI. It can help AI researchers, practitioners, and policy-makers to understand the XAI and IAI issues in the context of EAI and TAI.

    \item We offer the logic of outward and inward reasoning as the innovative approach to differentiate the XAI and IAI with various empirical experiments. This approach mitigates the confusion. It eliminates slippery, arbitrary, and subjective definitions of XAI and IAI. 

    \item This study demonstrates how to harness HPC or cloud power to fine-tune ML models based on a given dataset and employ well-established explainable techniques to explain ML models within an interpretable mindset.

    \item This work proposes a 3X3 high-level abstraction matrix for the meta-hyperparameter concept to define XAI’s criteria from data to ML modelling and from ML modelling to XAI across problems, hypothesis and validation/justification spaces. The search processing is determined by both a given dataset and mindset. The study illustrated the end-to-end (E2E) process of XAI and IAI experiments and highlighted details of the XAI/IAI pipeline.
    
\end{itemize}

The rest of the paper is organized as follows: \textbf{Section 2} provides a quick survey of eight well-known XAI techniques. \textbf{Section 3} introduces an open dataset for car insurance claims and our mindset regarding how to implement experiments at a high-level abstraction. \textbf{Section 4} presents the experimental results and further optimizes ML models through a hyperparameter search. \textbf{Section 5} discusses the results and gives a detailed analysis. \textbf{Section 6} outlines the conclusions and highlights future research works.

\section{Literature Review}
\subsection{Overview of XAI Techniques}
\label{subsec1}

\begin{figure}[htb]
    \centering
    \includegraphics[width= 1 \linewidth]{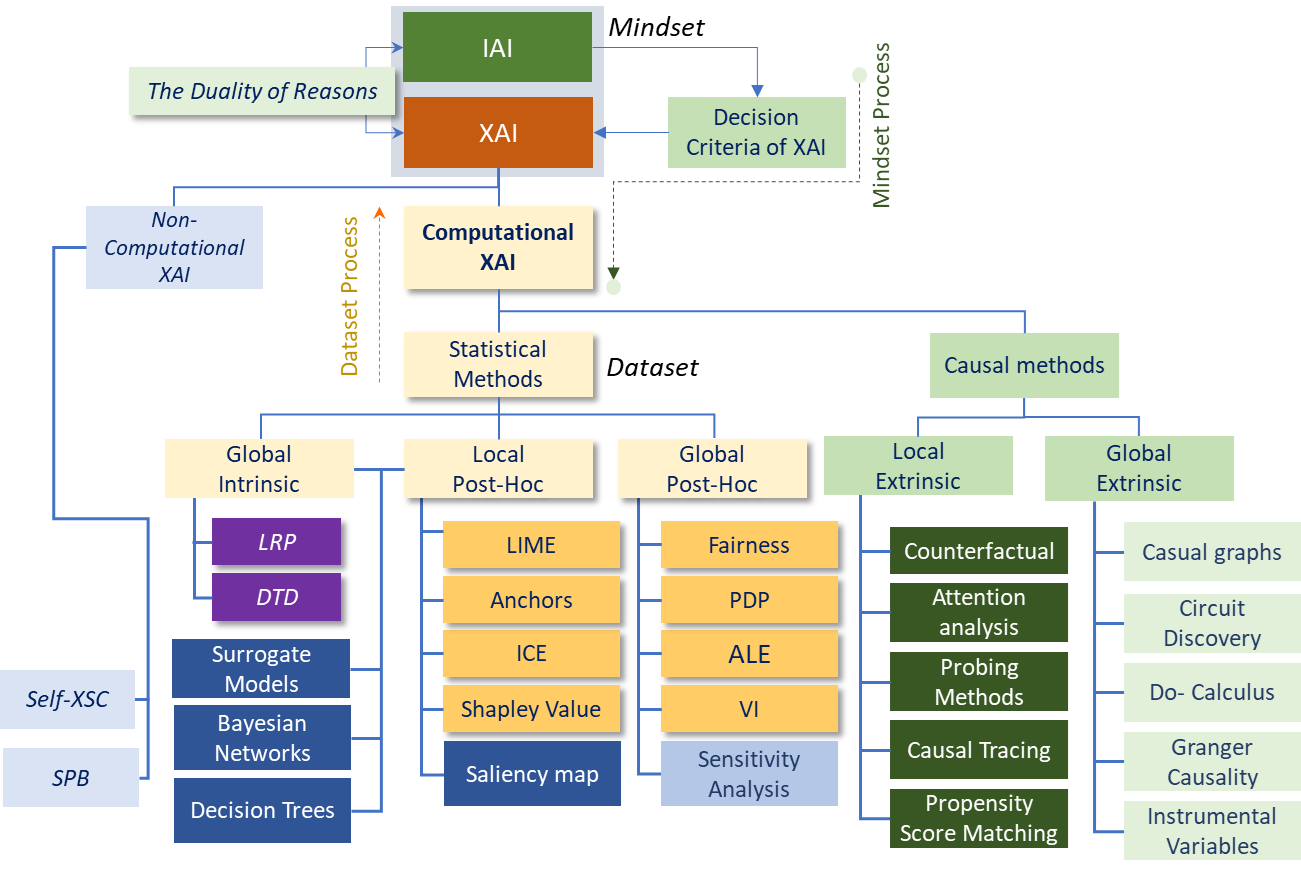}
    \caption[]{Taxonomy of XAI and IAI \footnote[1]{}}
    \label{fig:3}
    \vspace{-1 cm}
\end{figure}

Many widely recognized XAI techniques can be classified as computational versus non-computational, statistical methods versus causal methods (under computational), global versus local (under statistical), and post-hoc versus a priori(Refer to Fig ~\ref{fig:3}) if we assume XAI to be a subset of IAI from a top-down perspective. The meaning of "post-hoc" is to examine the result of the ML model after training \cite{Molnar16} because we want to understand how the final ML model reaches its conclusions. Due to limited space, we only concentrate on eight XAI techniques marked in orange. We exclude the saliency map and sensitivity analysis from this study because we want to focus on the gradient-boosting machines (GBM) as an ML model. Besides the global (model-agnostic) post-hoc, we can also have global intrinsic models, such as LRP \cite{Bach29} and DTD \cite{Montavon30} for neural networks. The meaning of intrinsic implies the model itself or opening a black box. In contrast, we can have extrinsic or causal models that can also be categorized locally and globally. Under the hood of the non-computational category, there are self-explanation scorecard (self-XSC) and stakeholder playbook (SPB) \cite{Hoffman32}. These XAI models are beyond the paper's scope.

\subsection{XAI Techniques Briefing}
The idea of Local Interpretable Model-agnostic Explanations or \textbf{LIME} \cite{Ribeiro11} is to employ a local and simple surrogate model($f_{s}$) to replace the original and complex model ($f_{M}$) known as local model-agnostic. The following equation can compute LIME:

\begin{eqnarray}\label{eq:vcg}
\xi(x) & = &
\arg\ \underset{f_{S}\in F}{min} \mathcal{L} (f_{M}, f_{S}, \pi_{x}) + \Omega(f_{S})
\end{eqnarray}

Where $\xi(x)$ is the error between $f_{M}$ (ML or target model) and $f_{S}$ (surrogate model) \textit{F} is a class of potential explainable models, which $f_{S} \in F$, $\Omega(f_{S})$ is the measurement of the complexity of $f_{M}$. $\pi_{x}$ defines a proximity measurement between an instance \textbf{z} and \textbf{x} as a locality around \textbf{x}. 

LIME is a very popular XAI tool. We can use it even as a debugging tool because it is simple and transparent. It can also be applied to many ML models. However, the quality of LIME depends on a surrogate model. The explainable results are often unstable. It gives rise to the Anchor model. 

\let\thefootnote\relax
\footnote{[1]\textbf{VI}:Variable Importance, \textbf{PDP}: Partial Dependent Plot, \textbf{ALE}:Accumulated Local Effects, \textbf{ICE}: Individual Conditional Expectation, \textbf{LRP}: Layer-wise Relevance Propagation, \textbf{DTD}: Deep Taylor Decomposition, \textbf{LRP}: Layer-wise Relevance Propagation, \textbf{DTD}: Deep Taylor Decomposition, \textbf{Self-XSC}: Self Explanation Score Card, \textbf{SPB}: Stakeholder Playbook}

\textbf{Anchor} is another local model-agnostic method presented by \cite{Wand12}\cite{Ribeiro13}. It aims to explain any black box model with high probability guarantees. The basic idea is to employ a decision rule (IF-THEN) for one or some instances while generalizing the rest. It does not aim to open the black box's architecture and understand the internal parameters of the model. Instead, it uses a similar approach as LIME to use high-precision rules called anchors for the target model. Thus, the anchor algorithm is universal and defined as follows:

\vspace{-0.3 cm}
\begin{align}
   {\mathbb{E}}_{\mathcal{D}(z|A)}[1_{f(x)=f(z)}]
   \geq{{\tau}, 
   A(x)} =1 
\end{align}

Where $x$ means the instance being explained, $A(x)$: Anchor is a set of predictions. If $A(x) =1 $ when $A$ is a sufficient condition for $f(x)$ with a high probability. The function $f(x)$ implies the classification model to be explained ${\mathcal{D}(\cdot|A)}$ is the distribution of neighbours of $x$ matching $\textbf{A}$. $\tau$ specifies a precision threshold. 

Comparing \textbf{LIME}, the \textbf{Anchor} approach provides the following \textbf{advantages:} 1.) The result is easier to comprehend due to IF-THEN rules; 2.) It can be applied to any model. It is less likely to underfit; 3.) It can also serve as a subset and cover some essential instances; 4.) It supports parallel computation. The \textbf{disadvantages} are 1.) The result heavily depends on the initial configuration; 2.) An outcome is too specific with low coverage; 3.) Building an anchor depends on many factors. It leads to the particular anchor of runtime varying widely; 4.) The explainable coverage is undefined in some domains. 

In order to achieve the total coverage, we can select the Individual Conditional Expectations (\textbf{ICE}) curve showing an ML model profile because the ICE plot gives the output of N instances \cite{Goldstein18} for features. The ICE plot provides heterogeneous relationships. The disadvantage of the ICE plot is that it can only show one feature at a time. To overcome the limitation, we can leverage the \textbf{Shapley value} method \cite{Frye27}. The concept of this approach has its roots in game theory, particularly cooperative game theory \cite{Fryer28}. The Shapley value only describes a very particular type of \textbf{fairness} that is equal based on the principle that all players deserve equal rights and opportunities.

To calculate a model's \textbf{fairness}, we can employ algorithmic fairness, which is one of the global techniques. It was proposed by Kozodoi et al. \cite{Kozodoi19} with 11 fairness data metrics, including demographic, proportional, equalized odds, predictive rate, false positive, false negative rate, accuracy, negative predictive value, etc. Given the limited space constraints, we only selected some metrics, such as accurate, precise, and predictive probability measurements. The advantages of accuracy and precision are that the prediction result can be intuitively simple. It improves the prediction results globally. However, it can be misleading in rare instances if the population size is immensely large, which explains the fairness regarding the population size.

Another global technique is the Partial Dependent Profile or Plot (\textbf{PDP}). It can demonstrate one or two features in the feature set \textbf{S} contributing to the final prediction result in marginal effects. The marginal effects mean that other features are excluded from the plot. The \textbf{PDP} can unveil the relationship \cite{Friedman15} between the input feature and the output prediction, whether linear, monotonic, or complex. The mathematical relationship \cite{Jerome31} can be expressed as follows:
\begin{align}
\hat{f_S} (x_S) = 
{\mathbb{E}}_{x_C}[\hat{f}(x_S, x_C)] = 
\int \hat{f} (x_S, x_C) d \mathbb{P}(x_C)
\end{align}%

Where $x_S$ donates to a feature. $x_C$ stands for other features of the ML model $\hat{f}$. The partial function $\hat{f}_S $ is an estimated value resulting from the average training data. It is similar to the Monte Carlo simulation, which predicts possible outcomes from an uncertain event.

The advantages of the \textbf{PDP} plot are straightforward and easier to implement. If the selected feature is not correlated, the PDP can represent how this feature impacts the predicted result on average. The disadvantages are: 1.) the maximum number of features is two; 2.) Some PDPs do not include feature distribution, which can cause misinterpretation. 3.) If the features in \textbf{C} and S are correlated, the PDP result cannot be trusted.

To solve the correlation issue, we can employ Accumulate Local Effect Plots (\textbf{ALE}). The ALE \cite{Apley17} uses conditional distribution plus the effect of correlated features of interest rather than marginal distribution. The ALE concept can be formulated in the following equations:
\begin{align}
f_{1, ALE}(x_1) = 
\int_{x_{min,1}} ^{x_1} {\mathbb{E}} [ f^1(X_1, X_2)|{X_1}= z_1 ] dz_2 - C_c \\ 
=\int_{x_{min,1}} ^{x_1} (p_{2|1} (x_2|z_1)f^1(z_1, x_2)dx_2)dz_1 - C_c
\end{align}%

Where $f^1(x_1, x_2)=\partial(x_1, x_2)/\partial{x_1}$ represent the local effect of $x_1$ on $f(\cdot)$ at $(x_1, x_2)$. $x_{min, 1}$ is the selected value near the lower bound of the effect support of $p_1(\cdot)$. $C_c$ is the constant chosen to centre the plot vertically $p_{2|1}({x_2}|{z_1})$ stands for conditional density probability. In essence, the \textbf{ALE} calculates the differences in predictions and only focuses on interesting features rather than all. The difference in predictions is a result of the feature's effects on all individual instances in a certain interval.

The advantages of the ALE are: 1) It still works if features are correlated; 2) The computation of the \textbf{ALE} is faster; 3) It is much easier and more transparent to explain results. The disadvantages of the ALE are: 1.) The result can be unstable at various intervals. There is no perfect solution to finding an ideal number of intervals; 2.) Unlike the PDP technique, we cannot use ICE as a complementary plot to check the ALE's heterogeneity in the feature effect; 3.) Although the second-order ALE estimates can present a changing stability across the feature space, it is not visible; 4.) Moreover, the second-order effect plot is hard to explain; 5.) Compared to the PDP, the ALE is much more complex in terms of implementation; 6.) The ALE can not exhibit more than two features.

The last technique is Variable Importance (VI) or Feature Importance (FI). It measures the importance of each variable or feature's contribution to the prediction result. We can permutate all variables or features and rank them in a list to visualize VI \cite{Breiman14}. Breiman proposed permutation VI for the random forests algorithm \cite{Molnar16}. The advantages of the VI are: 1.) It can be explained directly and efficiently; 2.) The VI provides an overview of all features. The disadvantages are 1.) If features are correlated, the interactive components between features cannot be added; 2.) The accuracy of the VI (or ranking order of VIs) heavily depends on the model's errors; and 3.) The VI is the result of the final model. All the above XAI techniques are also known as model-agnostic models that can be applied to any ML model.

\section{Database and Experimental Setup}
\subsection{Brief Overview of Dataset}

\begin{figure}[htb]
    \centering
    \includegraphics[width= 1 \linewidth]{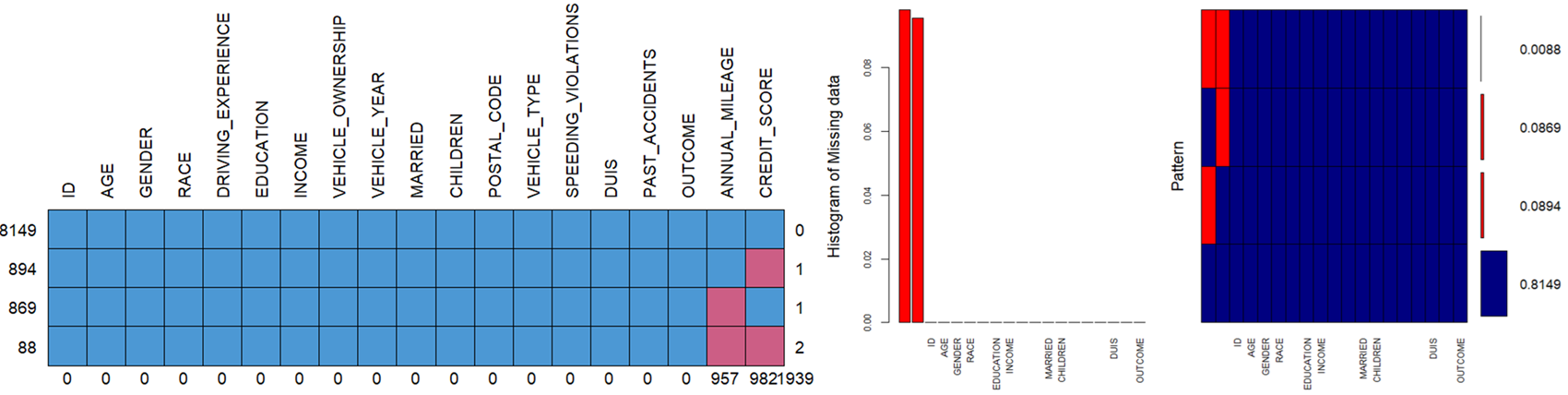}
    \caption{Dataset and Missing Values}
    \label{fig:4}
    \vspace{-.3 cm}
\end{figure}

Prior to a series of XAI experiments, it is necessary to understand a dataset. We selected an open dataset regarding annual car insurance claims from Kaggle \cite{Dataset20}. It comprises exactly 10,000 observations and 19 features. According to the data donor, most data is real, but the author changed some values. It is unclear which part of the data has been artificially modified. Furthermore, the dataset has 982 missing values in the "credit score" column and 957 missing values in the "annual mileage" column. Figure.\ref{fig:4} depicts the pattern of missing values. 

Although the missing values are slightly above a typical threshold level of 5\%, the previous experience \cite{Dong21} indicates that an imputation strategy is still better than simply omitting these missing values because 8.8\% is still within the principle guideline for imputation. Moreover, Madley-Dowd et al. \cite{Madley-Dowd22} argued that the proportion of missing data should not be used to guide decisions on multiple imputations. They prove that unbiased results can be achieved even with up to 90\% missing data. Considering the goal of this study, excluding missing values might diminish the explanatory power down the pipeline. Here is a critical decision point that differentiates between XAI and IAI. The decision to take what kind of technique depends on the individual's experience, intuition, and even belief or mindset. At first glance, data imputation seems to have nothing to do with the XAI model, but the data patterns will render the rules on our behalf. Consequently, selecting the right method of data imputation will directly impact XAI results. In other words, the dataset given mindset is equivalent to XAI given IAI. 

\subsection{Experimental Setup and Implementation}

Once the data imputation was completed, we split the dataset into a 70/30 ratio. The large part is for training, and the smaller proportion is for testing. We can also split the dataset into 80/20 or 50/50 ratios. It depends on a given problem context. Chollet\cite{Chollet37} suggested a 50/50 ratio for a time series problem because we try to predict the future given the past, not the reverse. We set the cross-validation value to five. 

The basic idea of the experiments is to train some common ML models first, such as general linear model (GLM), random forests (RF), gradient boost machine (GBM) and extended gradient boosting machine (Xgbm) that are considered to be a black box, and then we use explainable techniques, such as \textbf{VI, PDP, ICE, LIME, ALE, Shapley values, Anchors, and Fairness} to explain the black box for the prediction results.

We select the GBM to gain a basic understanding of the dataset. And then, we run Xgbm in parallel for a hyperparameter search to find an optimal solution. The solution will increase the accuracy of the prediction model. Usually, the optimal solution will take all features into account. Consequently, the Xgbm model becomes very challenging to explain. It leads to the immediate question: "Is the reason reasonable?" It also gives rise to the interpretation process. That is, when a reason turns to itself, it seeks to be justified and satisfied. We can somewhat perceive it as the reason reflecting on itself. It means the IAI process, which is iterative.

\section{Experiment Implementation, Results and Analysis}
\subsection{Prediction Results}

\begin{figure}[htb]
    \centering
    \includegraphics[width= 0.8 \linewidth]{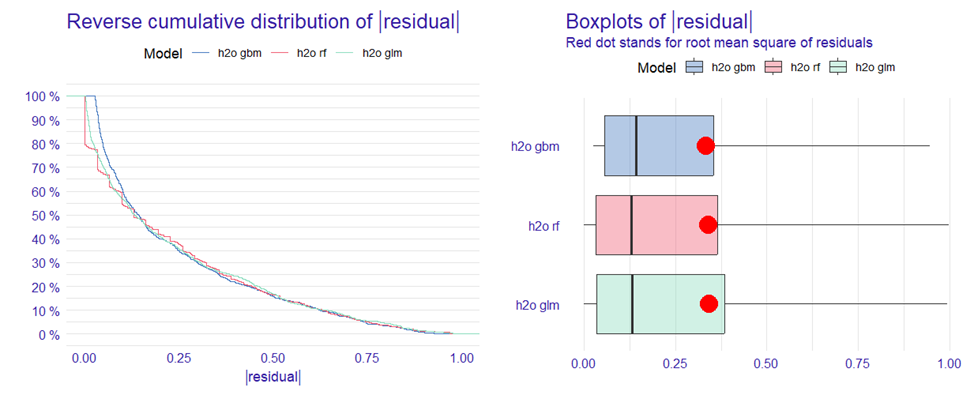}
    \caption{GLM RF and GBM Residuals Comparison}
    \label{fig:5}
    \vspace{-0.5 cm}
\end{figure}

The experiment results are divided into three folds: 1.) Prediction results via GLM\cite{Wood23}, RF, and GBM for comparison. 2.) Hyper-parameter searching for an optimizing solution via harnessing High-Performance Computing (or cloud computing) power. 3.) Explainable results by global and local XAI techniques. By comparison, GBM appears to be the best model to fit with the dataset.(See Figure ~\ref{fig:5}) However, the GBM result is not optimal. This leads to an optimization process via a hyperparameters search.

\subsection{Optimization Results}

The hyperparameter search begins with an initial random guess of the GBM model's parameters based on our intuition. These parameters include the number of trees, interaction depth, shrinkage rate, and etc. (See Table \ref{tab:1})

\begin{table} [htb]
\centering
\begin{tabular}{lllll}
\hline
Parameters       & Expt. 1 & Expt.2 & Expt.3 & Xgbm on HPC \\
\hline
number of Trees  & 5,000   & 10,000 & 12,000 & 1,000 \\
Learning Rate    & 0.001   & 0.01   & 0.1    & 0.05 \\
Tree Depth       & 1       & 3      & 5      & 5 \\
number min nodes & 1       & 3      & 2      & 1 \\
Col.Samples      &1        &1       &0.85    &0.9\\
Sub-sample       &1        &1       &0.75    &0.8\\
Non-Zero influ.  &12       &18      &18      &18 \\
CPU Time(Sec)    &6.86     &36.59   &65.53   &109,796.71 \\
Sys. time(Sec)   &0.05     &0.09    &0.19    &12.75 \\
E. Time(Sec)     &23.03    &93.46   &164.16  &865.85 \\
Best CV iter.    &5,000    &994     &84      &88\\
RMSE             &0.354    &0.325   &0.323   &0.312\\
\hline
\end{tabular}
\caption{Hyperparameter Search Configuration}
\label{tab:1}
\vspace{-0.3cm}
\end{table}

Table 1 shows that increasing the tree's number has minimal impact on the model performance. Decreasing the learning rate is not helpful either. Based on the initial trial, the Xgbm model was applied. We harness the HPC (or a cloud platform) for our hyperparameter search. The configuration of a computational cluster is 128 CPU cores with 256 GRAM.

This experiment set up 576 hyper-grid search grid points for an optimal solution. Although the final result does not seem to improve Root Mean Square Error (RMSE) performance significantly for the particular dataset due to the size of the dataset, this experiment aims to show the necessary steps for the E2E XAI | IAI process. If using GBM, a 576 grid points search will take over 30.5 hours in computation time. With an HPC platform, it only takes about 14 minutes to create a dataset with 10,000 instances. If the dataset has 100,000 or million instances, the HPC or cloud computing power will make a significant difference. This optimization step is essential to achieve a better explanation, especially for the VI technique. 

\subsection{Explanation AI}

\begin{figure}[htb]
    \centering
    \includegraphics[width=.9 \linewidth]{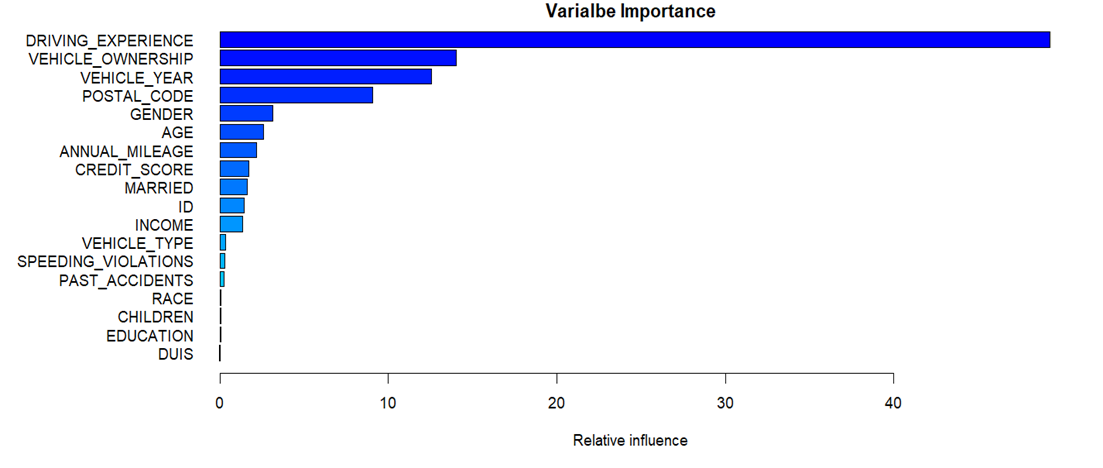}
    \caption{Variable Importance For Car Insurance Claim}
    \label{fig:6}
\end{figure}

\begin{figure}[htb]
    \centering
    \includegraphics[width=0.7\linewidth]{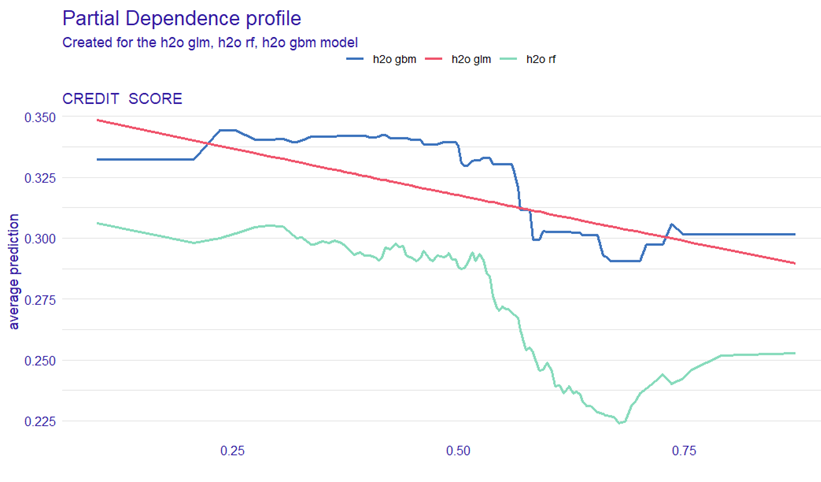}
    \caption{PDP Credit Score for three ML Models}
    \label{fig:7}
\end{figure}

\begin{figure}[htb]
    \centering
    \includegraphics[width=0.7\linewidth]{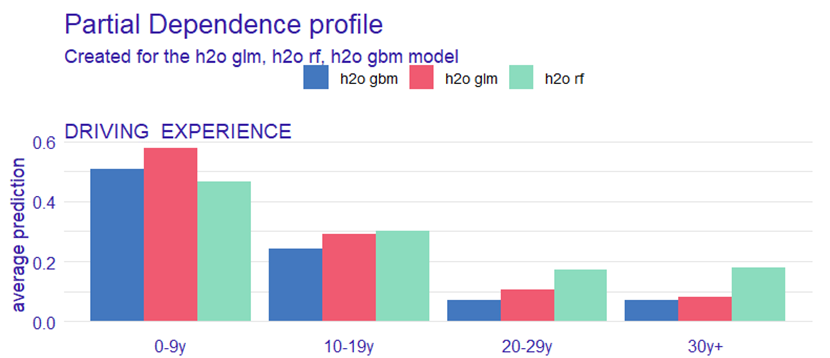}
    \caption{Average Prediction for Driving Experience}
    \label{fig:8}
    \vspace{-.3 cm}
\end{figure}

The first XAI experiment is the variable importance (VI) also known as feature influence (FI). It is very straightforward. Each feature gain can be calculated based on the optimal Xgbm model. (Figure ~\ref{fig:6}). Next, we can plot the PDP diagrams for the numerical variable (Credit Score) against the categorical variable (Driving experience). The PDP plot has three models [General Linear Model (GLM), Random Forest (RF), and Gradient Boost Machine (GBM)] for comparison. Figure. ~\ref{fig:7} illustrates the PDP result for the credit score feature. Figure. ~\ref{fig:8} displays the PDP result for the driving experience with three prediction models.

\begin{figure}[htb]
    \centering
    \includegraphics[width=1 \linewidth]{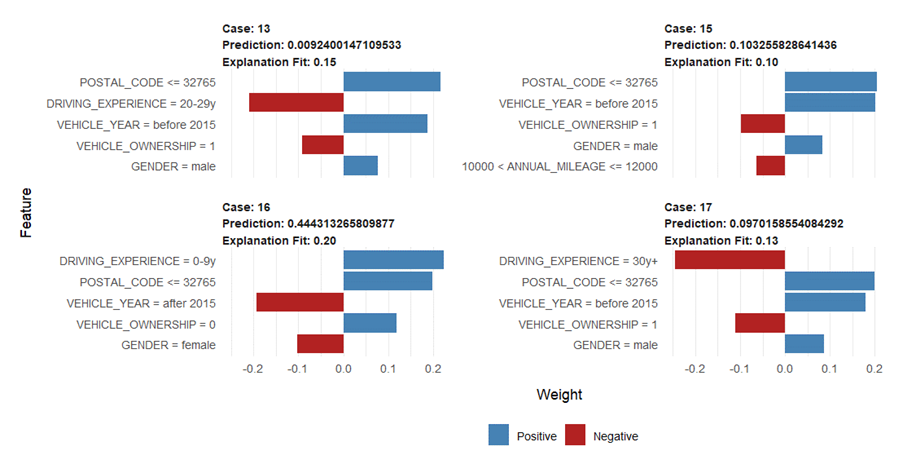}
    \caption{LIME XAI For Four Instances}
    \label{fig:9}
     \vspace{-.2 cm}
\end{figure}

To explain some particular instances, we can use the LIME technique to compute localized variable importance scores for a local profile because the LIME can often plot several instances simultaneously. Here, we randomly selected four instances to plot a local profile. (Refer to Figure ~\ref{fig:9})

As shown in Figure. ~\ref{fig:9}, some variables may contribute to the negative influence. For instance, the top left of Figure. ~\ref{fig:9} shows that if the driver's experience is between 20 and 29 years, it means the person has been rejected a car insurance claim.

\begin{figure}[htb]
    \centering
    \includegraphics[width=1\linewidth]{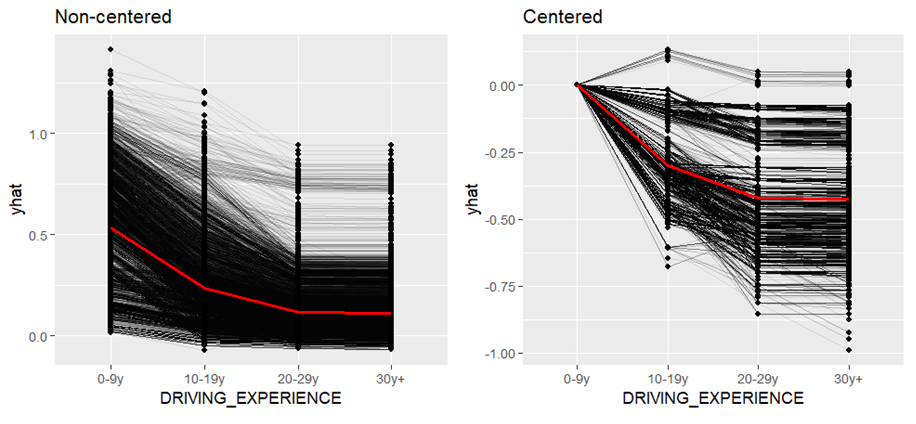}
    \caption{ICE Curve for Driving Experience}
    \label{fig:10}
    \vspace{-.2 cm}
\end{figure}

In addition to the LIME technique, the ICE is another local profile explanation. Figure. ~\ref{fig:10} displays the local profile for driving experiences. The ICE curves are extensions of the PDP. It plots each instance that contributes to the average predicted value.

Notice that the left diagram shows a stack style, while the right side exhibits the centred ICE. The centred ICE aims to visualize the heterogeneity in the prediction results. The decision on which diagram to adopt for explanation is dependent on the interpretation and even presentation.

\begin{figure}[htb]
    \centering
    \includegraphics[width=1.0\linewidth]{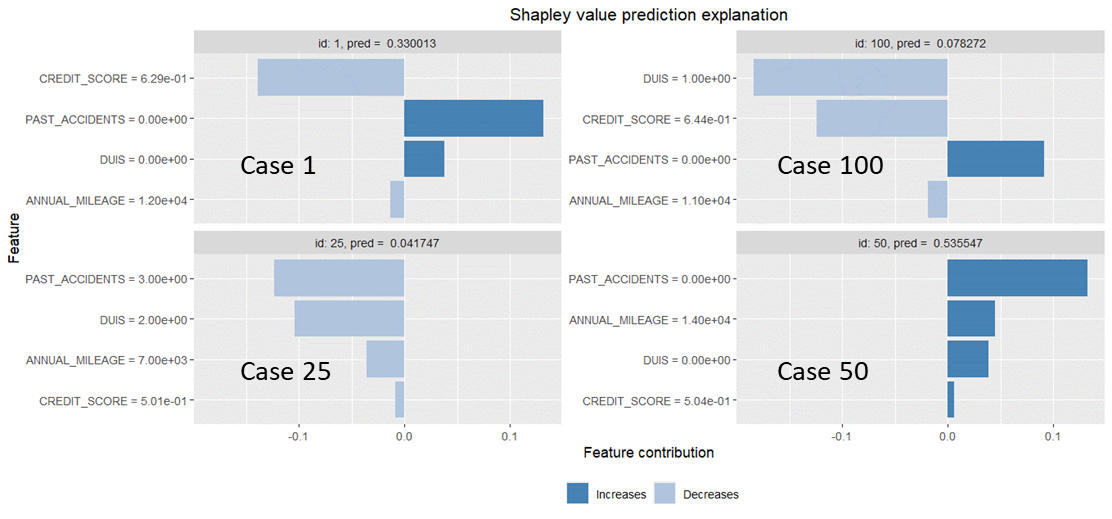}
    \caption{Shapley Value Explanation Cases: [1, 25, 50, 100]}
    \label{fig:11}
    \vspace{-.3 cm}
\end{figure}

Compared to LIME, the Shapley value evaluation can effectively work with numerical variables. Consequently, we select four numerical variables: "credit score", "past accident", "Driving Under the Influence (DUIS)", and "annual mileage". Then, we arbitrarily select four instances or observations [1, 25, 50 and 100] for the experiment. The final Shapley value plot shows four observations [1, 25, 50, 100] in Figure ~\ref{fig:11}. They are different, especially cases 25 and 50, which are opposite regarding explainable features.

\begin{figure}[htb]
    \centering
    \includegraphics[width=1 \linewidth]{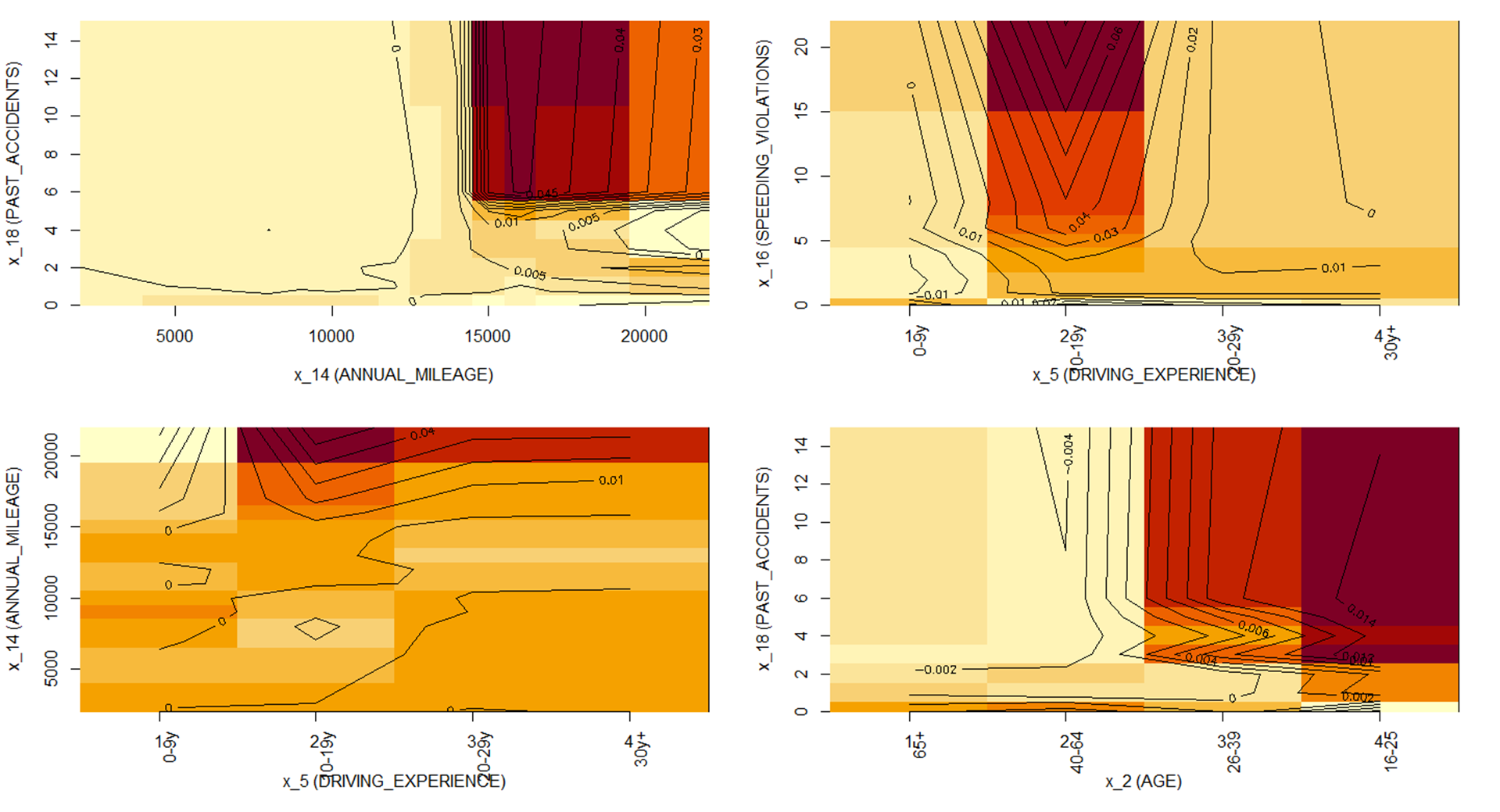}
    \caption{ICE Curve for Driving Experience (left) and Age (Right)}
    \label{fig:12}
    \vspace{-.2 cm}
\end{figure}

All the above experiments assume that features are not interactive with each other. However, this assumption could face a challenge. For example, the features of past accidents and annual mileage could be correlated because, intuitively, the higher the driving mileage, the more accidents could occur. We can verify our intuition or hypothesis with the ALE plot. Figure ~\ref{fig:12} has approved that our intuition is right but only after the threshold level of annual mileage between 13,000 and 14,000. (Refer to Figure. ~\ref{fig:12}, at left top)

Figure. ~\ref{fig:12} demonstrates four pairs of features (past accident versus annual mileage, speeding violations versus driving experience, annual mileage versus driving experience, and past accidents versus age) regarding the correlation of the features. Some features are closely correlated if the value is within a specific range. For example, the categorical variable of age and the numerical variable of past accidents are closely correlated if the ages are between 16 and 25. (see Figure ~\ref{fig:12} at the bottom right). The decision of selecting which diagram to explain may rely on the researcher's interpretation and expertise. For this particular dataset, we only have 19 features. If the dataset has 100 or even 1000 features, the number of feature combinations could be significantly large. Consequently, we need a high-level abstraction or IAI to subjectively select the desired feature combination for XAI. It is apriori. That is, we want reason to be justified.

\begin{figure}[htb]
    \centering
    \includegraphics[width=0.85\linewidth]{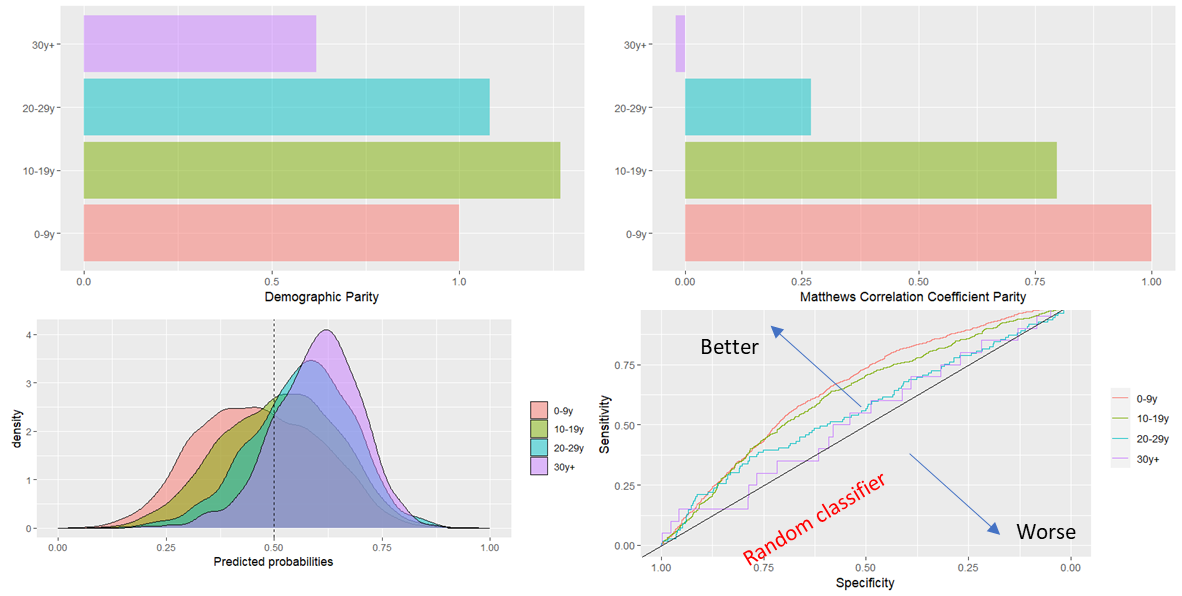}
    \caption{Fairness: Demographic and Predictive Probability}
    \label{fig:13}
    \vspace{-.3 cm}
\end{figure}

Another instance of IAI is fairness, a concept introduced by Mitchell\cite{Mitchell24} and Friedler\cite{Friedler25}. The core of this fairness is to ensure that the prediction result does not discriminate against underrepresented subgroups in the dataset. In statistics, this is known as sensitivity (true positive rate: TPR) and specificity (true negative rate: TNR). The 'demographic parity' metric, which is the sum of TRP + TNR, is particularly crucial as it provides an objective measure of fairness in the prediction results.Figure.~\ref{fig:13} presents Mathews Correlation Coefficient (MCC) or Phi coefficient ($\varphi$) (at the top right) and Receiver Operating Characteristic (ROC) results (at the bottom right). These two metrics demonstrate that predicting the OUTCOME for individuals with over 30 years of driving experience is almost impossible, as the ROC metric (purple colour) is very close to the diagonal line, indicating randomness. The MCC measurement further confirms that the binary of OUTCOME is not correlated with driving experience over 30 years. Conversely, the binary variable OUTCOME is closely correlated for individuals with 0-9 years of driving experience, providing an objective measure of fairness.

\begin{figure}[htb]
    \centering
    \includegraphics[width= 1 \linewidth]{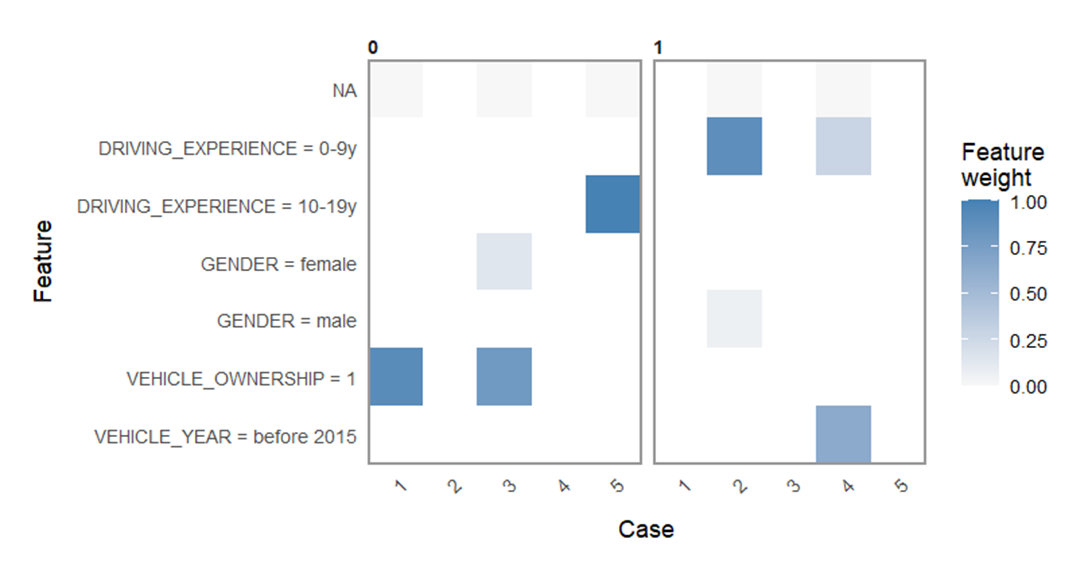}
    \caption{Anchors Explainable AI by Five Instance}
    \label{fig:14}
\end{figure}

The last XAI experiment is the "anchors" technique, which uses a simple decision tree (IF-THEN) to explain the ML model. It means that the anchors use one or a few particular instances (or anchors) with decision rules to explain the ML model while generalizing to as many other instances as possible. The experiment uses five instances (anchors) to explain the ML model. (Refer to Figure. ~\ref{fig:14} and Table \ref{tab:2}). Notice that the selected case 5 only has 32.5\% instances coverage, although the prediction precision is 100\%, while case 1 can achieve 90.8\% prediction with nearly 70\% coverage.  

\begin{table}[htb]
\centering
\begin{tabular}{rlllll}
\hline
Case       & 1 & 2 & 3 & 4 & 5 \\
\hline
Pred.Pre. &90.8\% &94.4\% &92.7\% &92.4\% &100.0\% \\
Pred.Cov. &69.9\% &15.6\% &34.0\% &30.8\% & 32.5\%\\
\hline
\end{tabular}
\caption{ANCHORS CASE}
\label{tab:2}
\end{table}

So far, we have implemented eight XAI experiments to demonstrate that many IAI decision points or high-level abstractions are required during the implementation of XAI experiments. There is a clear demarcation between XAI and IAI. We argue these two notions are totally different. We cannot use XAI and IAI interchangeably. Without IAI, XAI experiments could not be executed.

\section{Results Discussion}
\begin{figure}[htb]
    \centering
    \includegraphics[width= 1\linewidth]{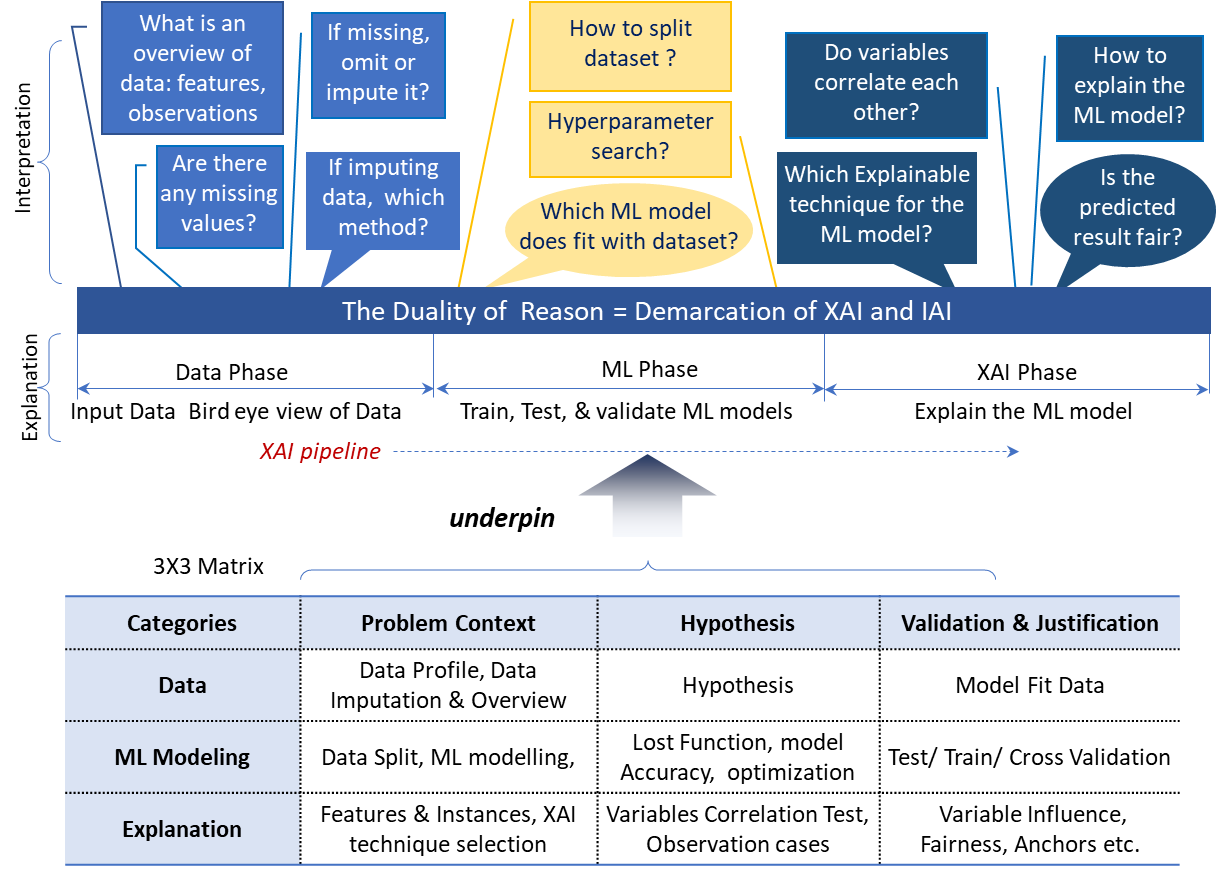}
    \caption{Three Phases of Experiment}
    \label{fig:15}
    \vspace{-.2 cm}
\end{figure}

The paper's primary research question is whether the XAI and IAI differ. If so, why is it so important? How can we distinguish between them? We solve these problems in three phases: data processing, ML modelling, and analysis of XAI techniques. Each phase has many subsequent questions or high-level abstractions for the goal of XAI. The approach is similar to Biecek and Burzykowski's \cite{Biecek26} explanatory method. We can summarize the high-level abstraction into a 3X3 matrix that consists of 1) data, 2) ML modelling, and 3) explanation multiplied by 1) problem context, 2) hypothesis, and 3) validation \& justification. (See Figure \ref{fig:15}). This high-level abstraction of XAI can form a meta-hyperparameter search mechanism that could be implemented iteratively. 

For the data processing phase, we performed a data imputation experiment (Refer to Figure. ~\ref{fig:4}). We believe the imputation data method is better than the omitted missing data approach for XAI. Our belief is a part of our intuition. We can ask many high-level reflected questions to interpret the given dataset for the ML problem. 

During the ML phase, three ML models are generated: GLM, RF, and GBM with its extension of GBM (or Xgbm) for comparison of prediction accuracy (See Figure. ~\ref{fig:5}). We can also adopt a transformer model (deep learning architecture) for prediction. The choice of which ML model to use depends on the expected result of the prediction and subsequent selection of XAI methods. (e.g. global intrinsic or global post-hoc). It determines our mindset for a given dataset. When we build an ML model to fit with a dataset, we must consider at least five issues for the future XAI: 1.) Which ML model should be selected for the context of the ML problem? 2.) What is the dataset ratio for training, testing, and validation? 3.) Which feature or features do I care about the most? 4.) How can the ML model be optimized? All these questions are within a framework of a meta-hyperparameter search that defines XAI criteria.

In general, the ML model may provide some clues for adopting which XAI technique. For example, LRP and DTD are usually applied to neural networks. Some XAI techniques are universal and can be applied to any type of black box, such as LIME. As we showed above, the VI and PDP plots are often unstable. The variables' ranking order often changes if the RMSE is drifting. It may indicate that some features are correlated. The ALE plot is an alternative to the PDP. We show four pairs of correlation features in Figure ~\ref{fig:12}. If we want to focus on the local explanation, LIME, ICE, Shapley value, and Anchor techniques offer solutions for XAI. However, the selection of which XAI technique depends on our mindset or IAI regarding the problem context. 

\section{Conclusion and Future Works}
We argue that XAI and IAI differ. The demarcation of XAI and IAI is the duality of the reason because whether we want to explain or interpret AI, we must provide some reasons. However, reason has its duality, also known as outward and inward reasoning. When we reason outwards, we want a reason to make sense, which is governed by the law of nature, logic, algorithms, rationality, and dataset. When we reason inwards, we want reason to be happy. It is governed by the law of the heart that eventually leads to ethics, belief, and intuition or mindset. Our E2E explanation process demonstrates that many decision points or criteria of XAI are required based on our mindset and interaction with the XAI process. From a programming perspective, IAI is similar to one level-up of abstraction, while XAI is more like detailed commands that can get things done.  

The implications of this demarcation clarify the notions between XAI and IAI. This clarification can help many practitioners and policymakers move beyond simple algorithmic explanations. This work's main contribution highlights the demarcation through various empirical experiments. 

The limitation is that we are unable to test deep neural networks-related XAI tools, such as LRP and DTD, as well as the causal methods for XAI. Moreover, how can we interpret the explainable results when we apply the global intrinsic approach for explanation? Is there a Gestalt shift for the intrinsic approaches? How can we generate high-level abstraction for a meta-hyperparameter search? These are our future works. We will explore these questions with other XAI techniques to understand how the other XAI techniques impact IAI.

\section{Acknowledge}

This research was supported by the Luxembourg National Research Fund (FNR), and funding was provided by grant ID C21-IS-16221483-CBD and grant ID 15748747.





\begin{thebibliography}{00}


\bibitem{Guidotti1}
    Riccardo Guidotti,
    \textit{A survey of methods for explaining black box models. ACM computing surveys (CSUR)}
    ACM COMPUTING SURVEYS, Vol. 51.5, p1-42
    August
    2018.

\bibitem{Miller2}
    Tim Miller,
    \textit{Explanation in artificial intelligence: Insights from the social sciences},
    Artificial intelligence, Vol. 267, p1-38
    February
    2019

\bibitem{Lipton3}
    Zachary C. Lipton,
    \textit{The mythos of model interpretability: In machine learning, the concept of interpretability is both important and slippery}
    Queue, Vol.16.3, p.31-57
    June
    2018

\bibitem{Benois4}
    Jenny Benois-Pineau(Ed),
    \textit{Explainable Deep Learning AI: Methods and Challenges}
    1st, Elsevier, Science,
    Oxford UK
    February
    2023

\bibitem{Biran5}
    OrBiran and Courtenay Cotton,
    \textit{Explanation and justification in machine learning: A Survey}
    IJCAI-17 workshop on explainable AI, Vol.8.1. 201, p8-13
    August
    2017

\bibitem{Burkart6}
    Nadia Burkart and Marco F. Huber,
    \textit{A survey on the explainability of supervised machine learning} 
    Journal of Artificial Intelligence Research,
    vol.70,
    p245-317
    Jan 
    2021

\bibitem{DARPA7}
    DARPA
    \textit{Performance Maximisation}
    DARPA report
    June
    Access 2023
    https://www.darpa.mil/attachments/DARPA-BAA-16-53.pdf,
    
\bibitem{Zhong8}
    Jinfeng Zhong and Elsa Negre,
    \textit{AI: To interpret or to explain?}
    INFormatique des ORganisations et Systèmes d'Information et de Décision,
    June
    Access 2023,
    https://inforsid2023.sciencesconf.org/

\bibitem{Oneill9}
    Onora O'neill,
    \textit{Constructions of Reason: Explorations of Kant's practical philosophy}
    1st,
    Cambridge University Press
    January
    1989

\bibitem{Logins10}
    Artūrs Logins
    \textit{Normative Reasons: Between Reasoning and Explanation}
    1st,
    Cambridge University Press
    August
    2022

\bibitem{Ribeiro11}
    Marco Tulio Ribeiro
    \textit{Why should I trust you? Explaining the predictions of any classifier}
    In Proceedings of the 22nd ACM SIGKDD international conference on knowledge discovery and data mining,
    vol.1,
    p1135-1144,
    August
    2016

\bibitem{Wand12}
    Jonathan Wand,
    \textit{Anchors: Software for anchoring vignette data}
    Journal of Statistical Software
    vol.42,
    p1-25
    June
    2011

\bibitem{Ribeiro13}
    Marco Tulio Ribeiro,
    \textit{Anchors: high-precision model-agnostic explanations}
    In: Proceedings of the AAAI Conference on Artificial Intelligence,
    vol.32.1,
    p11491
    April
    2018

\bibitem{Breiman14}
    Leo Breiman
    \textit{Random forests}
    Machine learning
    vol.45,
    p5-32
    October
    2001

\bibitem{Friedman15}
    Jerome H. Friedman,
    \textit{Greedy function approximation: a gradient boosting machine}
    Annals of Statistics
    vol.1,
    p1189-1232
    October
    2001

\bibitem{Molnar16}
    Christoph Molnar,
    \textit{Interpretable machine learning}
    1st
    Lulu.com
    p116-120
    February
    2020

\bibitem{Apley17}
    D.W. Apley and Jingyu Zhu,
    \textit{Visualizing the effects of predictor variables in black box supervised learning models}
    Journal of the Royal Statistical Society Series B: Statistical Methodology,
    vol.82.4
    p1059-1086
    September
    2020

\bibitem{Goldstein18}
    Alex Goldstein,
    \textit{Peeking inside the black box: Visualizing statistical learning with plots of individual conditional expectation}
    Journal of Computational and Graphical Statistics,
    vol.24.1,
    p44-65
    January 
    2015

\bibitem{Kozodoi19}
    Nikita Kozodoi,
    \textit{Fairness in credit scoring: Assessment, implementation and profit implications}
    European Journal of Operational Research
    vol.297.3,
    p1083-1094
    March
    2020
    
\bibitem{Dataset20}
    Kaggle
    \textit{Car insurance dataset}
    Access 2023
    \url{https://www.kaggle.com/datasets/sagnik1511/car-insurance-data}
    June
    2023

\bibitem{Dong21}
    Yiran Dong
    \textit{Principled missing data methods for researchers}
    SpringerPlus
    vol.2,
    p1-17
    December
    2013

\bibitem{Madley-Dowd22}
    Paul Madley-Dowd
    \textit{The proportion of missing data should not be used to guide decisions on multiple imputation}
    Journal of Clinical Epidemiology
    vol.110
    p63-73
    June
    2019

\bibitem{Wood23}
    Simon. N Wood
    \textit{Generalized additive models: an introduction with R}
    1st
    CRC Press
    May
    2017

\bibitem{Mitchell24}
    Shira Mitchell
    \textit{Algorithmic fairness: Choices, assumptions, and definitions}
    Annual Review of Statistics and Its Application
    vol.8
    p141-163
    March
    2021

\bibitem{Friedler25}
    Sorelle A. Friedler
    \textit{A comparative study of fairness-enhancing interventions in machine learning}
    Proceedings of the conference on fairness, accountability, and transparency
    vol.329
    p329-338
    January
    2019

\bibitem{Biecek26}
    Przemyslaw Biecek and Tomasz Burzykowski
    \textit{Explanatory model analysis: explore, explain, and examine predictive models}
    1st
    CRC Press
    September
    2021

\bibitem{Frye27}
    Christopher Frye
    \textit{Shapley explainability on the data manifold}
    arXiv preprint
    arXiv:2006.01272
    https://arxiv.org/abs/2006.01272 
    June
    2020

\bibitem{Fryer28}
    Daniel Fryer
    \textit{Shapley values for feature selection: The good, the bad, and the axioms}
    IEEE Access
    vol.9
    p144352-144360
    October
    2018

\bibitem{Bach29}
    Sebastian Bach
    \textit{On pixel-wise explanations for nonlinear classifier decisions by layer-wise relevance propagation}
    PloS one
    vol.10.7
    pe0130140,
    July
    2015

\bibitem{Montavon30}
    Grégoire Montavon
    \textit{Explaining nonlinear classification decisions with deep Taylor decomposition}
    Pattern recognition
    vol.65
    p211-222
    May
    2017

\bibitem{Jerome31}
    Jerome H. Friedman
    \textit{Greedy function approximation: a gradient boosting machine}
    Annals of Statistics
    vol.1,
    p1189-1232
    October
    2001

\bibitem{Hoffman32}
    Hoffman, Robert R., et al.
    \textit{Explainable AI: roles and stakeholders, desirements and challenges}
    Frontiers in Computer Science
    Vol.5
    p1117848
    August
    2023

\bibitem{Mohar33}
    Molnar, Christoph. 
    \textit{Modeling Mindsets: The Many Cultures of Learning from Data} 
    1st Edition
    Mucbook, 
    November
    2022.

\bibitem{Greenwell34}
    Brandon M. Greenwell
    \textit{Tree-Based Methods for Statistical Learning in R}
    1st Edition
    CRC Press
    p203-357
    June
    2022

\bibitem{Dandl35}
    Susanne Dandl et al.
    \textit{Applied Machine Learning Using mlr3 in R}
    1st Edition
    CRC Press
    January 
    2024
     
\bibitem{bradley36}
    Bradley Boehmke and Brandon Greenwell
    \textit{Hands-on machine learning with R}
    1st Edition
    https://uc-r.github.io/dalex
    CRC Press
    November
    2019

\bibitem{Chollet37}
    Francois Chollet, et al.
    \textit{Deep Learning with R}
    2nd Edition
    Manning Publications Co
    p301-333
    July
    2022

\bibitem{Popova38}
    Zhuhadar, Lily Popova, and Miltiadis D. Lytras.
    \textit{The application of AutoML techniques in diabetes diagnosis: current approaches, performance, and future directions}
    Sustainability
    Vol.15.18
    May
    2023

    
\end{thebibliography}



\end{document}